\pdfoutput=1

\documentclass[11pt]{article}

\usepackage[]{ACL2023}

\usepackage{times}
\usepackage{latexsym}
\usepackage[most]{tcolorbox}

\usepackage[T1]{fontenc}

\usepackage[utf8]{inputenc}

\usepackage{microtype}

\usepackage{inconsolata}
\usepackage{bibentry}
\usepackage{url}
\usepackage{booktabs}
\usepackage{enumitem}
\usepackage{caption}
\usepackage{subcaption}
\usepackage{soul}
\usepackage{multirow}
\usepackage{amsmath}
\usepackage{makecell}
\usepackage{fdsymbol}

\usepackage{graphicx}
\usepackage{subcaption}
\usepackage{xcolor}
\usepackage{balance}

\newcommand{\gtlogo}{\raisebox{3.4pt}{\includegraphics[scale=0.015]{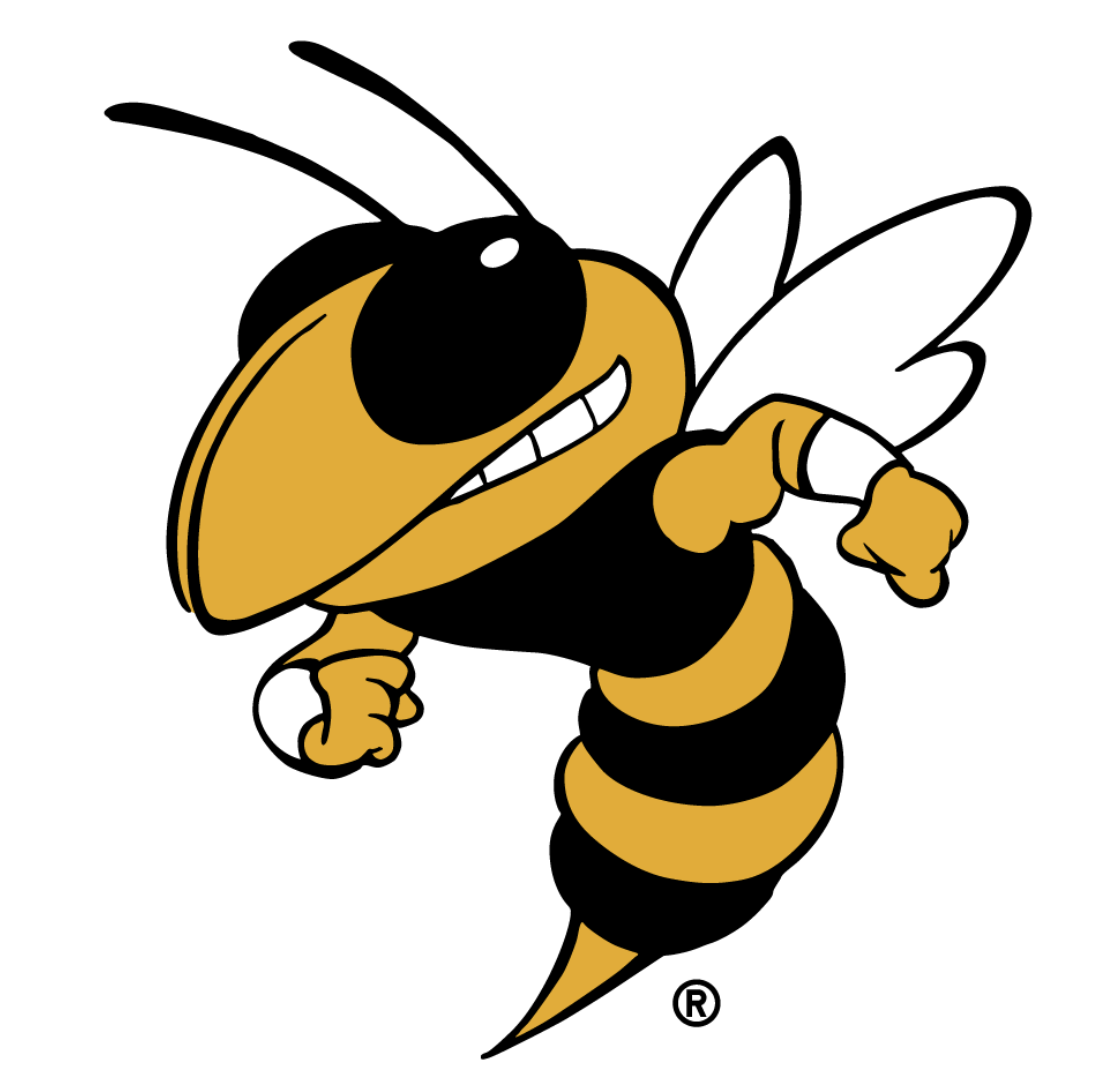}}}
\newcommand{\hulogo}{\raisebox{3.4pt}{\includegraphics[scale=0.005]{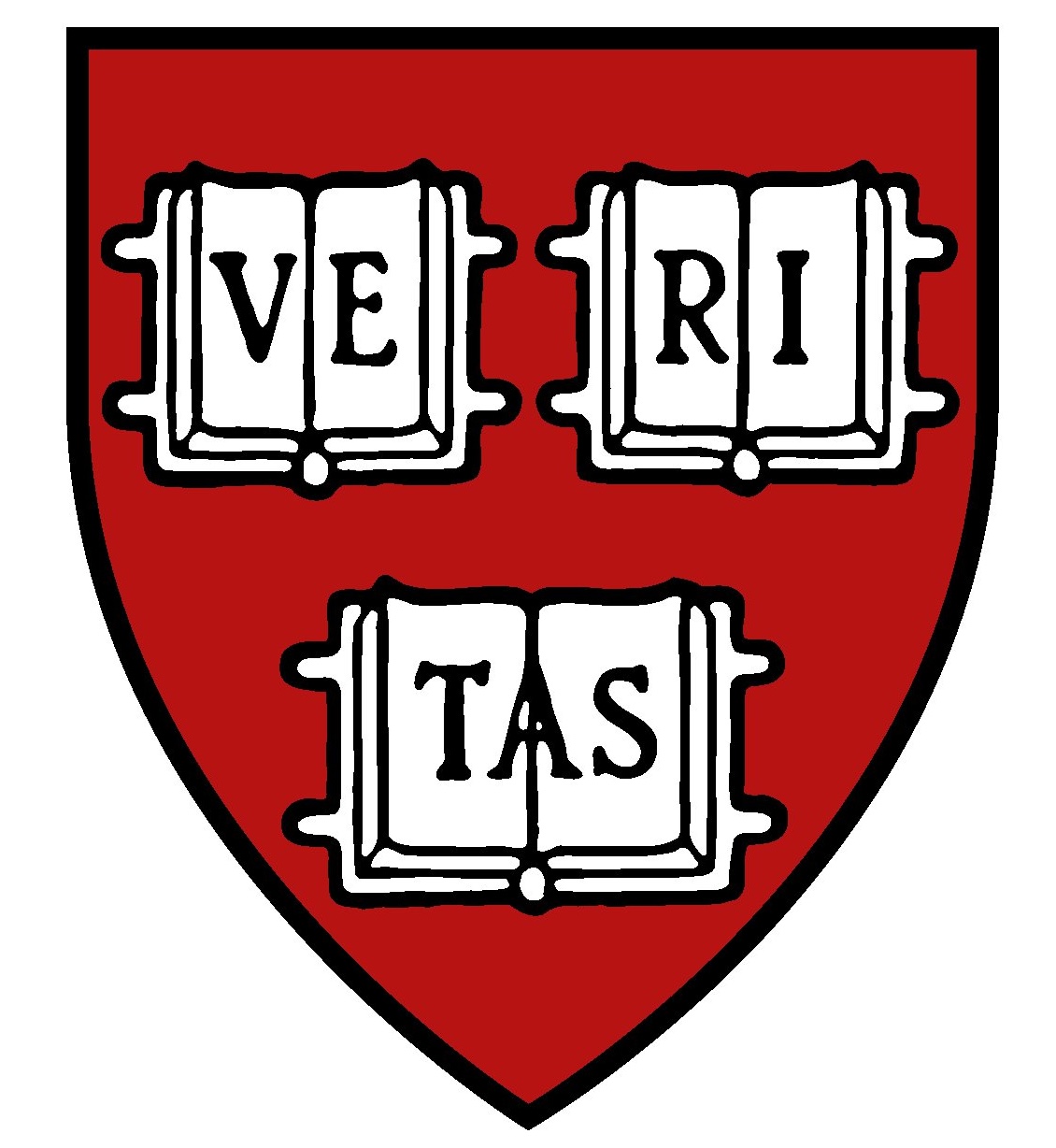}}}
\newcommand{\gt}{\gtlogo}
\newcommand{\hu}{\hulogo}

\definecolor{bondiblue}{rgb}{0.0, 0.58, 0.71}
\definecolor{inc}{rgb}{0.09, 0.45, 0.27}
\definecolor{dec}{rgb}{0.66, 0.13, 0.24}

\title{Cross-Modal Projection in Multimodal LLMs Doesn’t Really Project\\ Visual Attributes to Textual Space}

\author{Gaurav Verma \gt \hspace{1.5em}
        Minje Choi \gt \hspace{1.5em}
        Kartik Sharma \gt \\
        \textbf{Jamelle Watson-Daniels} \hu \hspace{1.5em}
        \textbf{Sejoon Oh} \gt \hspace{1.5em}
        \textbf{Srijan Kumar} \gt \hspace{1.5em} \\
        \gt Georgia Institute of Technology, \hu Harvard University\\
        \texttt{\{gverma, minje.choi, ksartik, soh337, srijan\}@gatech.edu}\\
        \texttt{jwatsondaniels@g.harvard.edu}
        }
  
\begin{document}
\maketitle
\begin{abstract}

Multimodal large language models (MLLMs) like LLaVA and GPT-4(V) enable general-purpose conversations about images with the language modality. As off-the-shelf MLLMs may have limited capabilities on images from domains like dermatology and agriculture, they must be fine-tuned to unlock domain-specific applications. The prevalent architecture of current open-source MLLMs comprises two major modules: an image-language (cross-modal) projection network and a large language model. It is desirable to understand the roles of these two modules in modeling domain-specific visual attributes to inform the design of future models and streamline the interpretability efforts on the current models. To this end, via experiments on $4$ datasets and under 2 fine-tuning settings, we find that as the MLLM is fine-tuned, it indeed gains domain-specific visual capabilities, but the updates do \textit{not} lead to the projection extracting relevant domain-specific visual attributes. Our results indicate that the domain-specific visual attributes are modeled by the LLM, even when only the projection is fine-tuned. Through this study, we offer a potential reinterpretation of the role of cross-modal projections in MLLM architectures.
\end{abstract}

\section{Introduction}

\begin{figure}[t!]
    \centering
    \includegraphics[width=1.02\linewidth]{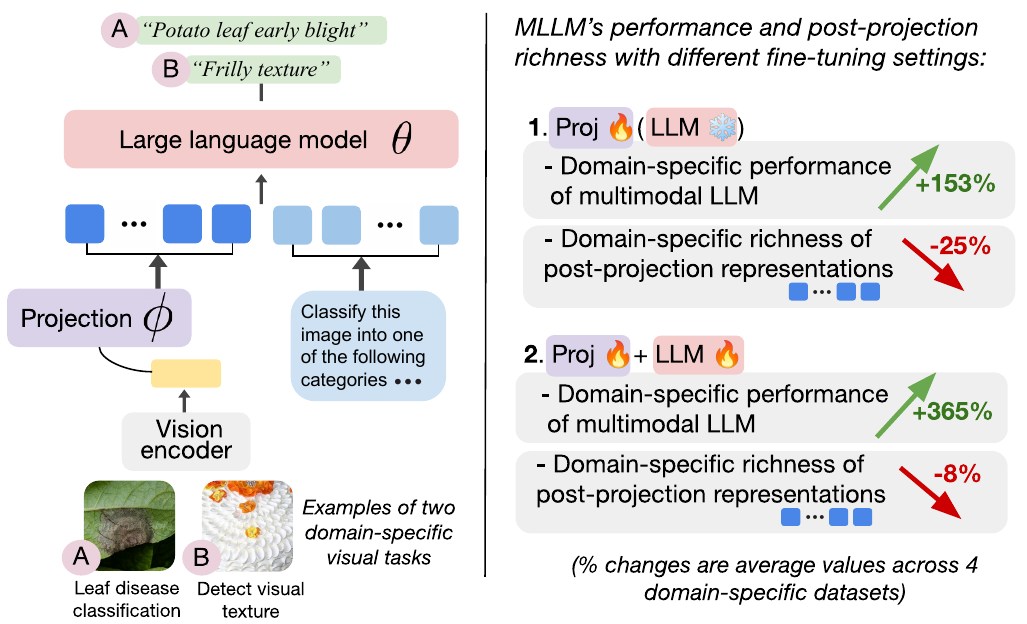}
    \caption{{\textbf{Overview of our study.} While the MLLM's domain-specific visual capability can be improved using fine-tuning strategies, the domain-specific richness of the image's post-projection representation does not improve. Results indicate that domain-specific visual attributes are predominantly modeled by the LLM parameters (whether frozen or not) and the projection does not necessarily play a role in mapping visual attributes to the LLM space. 
    }}
    \label{fig:overview}
\end{figure}

The recent wave of advancements in large language models (LLMs) has equipped them with the ability to ``see'' images, leading to multimodal large language models (MLLMs) like LLaVA~\cite{liu2023visual}, GPT-4(V)~\cite{achiam2023gpt}, and Gemini~\cite{team2023gemini}. MLLMs unlock the potential to converse with visual data using language. However, existing MLLMs are trained and evaluated for general-purpose multimodal tasks like question-answering on \textit{natural images}\footnote{We use `natural images' or `internet images' to refer to common images encountered on social media platforms and the Web and contrast them with domain-specific images.}~\cite{liu2023visual, LandingAI2024}, which limits their applicability in specific domains like agriculture and dermatology. MLLMs with domain-specific visual capabilities can transform workflows in several industries, including healthcare, agriculture, circuit design, and satellite imaging~\cite{miotto2018deep, ferentinos2018deep, anilturk2023automatic, kaselimi2022vision}. While fine-tuning can improve domain-specific visual capabilities of general-purpose MLLMs, we adopt domain-specific fine-tuning as a strategic approach to understand the roles that the MLLM's key architectural components play in modeling visual attributes. A better understanding of the roles of MLLM's components in modeling visual attributes can inform future design choices as well as direct interpretability efforts. 

Architecturally, open-source MLLMs comprise two key components: \textit{(i)} a cross-modal projection layer that connects image representations with the LLM, and \textit{(ii)} the LLM that processes the projected image representation and the text tokens; see Figure \ref{fig:overview} (left). In the context of the projection, researchers often consider the projection layer as the unit responsible for aligning features/concepts from the image to the LLM space~\cite{li2023llava, lin2023video, moon2023anymal}. Consequently, one prevalent fine-tuning strategy to adapt MLLMs for domain-specific visual tasks is to update the projection while keeping the LLM parameters frozen~\cite{moon2023anymal}. Alternatively, the projection and the LLM parameters can be fine-tuned concurrently~\cite{liu2023improved}. 

In this work, we use domain-specific fine-tuning using the above two strategies to understand the role of the projection and the LLM parameters in acquiring domain-specific image modeling capabilities. We posit that if the projection plays a critical role in acquiring domain-specific image modeling capabilities, the post-projection representation -- i.e., the representation of the image transformed by the projection, should be \textit{richer}\footnote{We use domain-specific richness to indicate the ``expressive power'' of the representations~\cite{bengio2012unsupervised} towards the domain-specific task.} in domain-specific features. 
Conversely, if the post-projection representation is not richer in domain-specific features, the domain-specific features are being identified or modeled by the LLM parameters.\footnote{Project webpage: \url{https://claws-lab.github.io/projection-in-MLLMs/}}

Our experiments and analysis with $4$ different datasets show that, as expected, both the fine-tuning strategies boost domain-specific closed-set image classification performance of the MLLM. However, none of the strategies lead to extraction of richer domain-specific features by the update in the projection layer; see Figure \ref{fig:overview} (right). This indicates that as MLLMs are fine-tuned to classify domain-specific images, the identification of domain-specific image attributes occurs in the LLM parameters, whether frozen or not. More broadly, our results add to the existing evidence that deep neural networks can be inherently multimodal~\cite{goh2021multimodal, schwettmann2023multimodal}, and LLMs could model visual data with minimal assistance from the cross-modal projection.

We first discuss the fine-tuning strategies to improve the domain-specific capabilities of MLLMs (Section \ref{sec:comparison}) and then analyze the role of projection in acquiring the new domain-specific capabilities (Section \ref{sec:hook_and_extract}). Finally, we discuss the implications of our work and the future directions (Section \ref{sec:discussion}).

\section{Effect of Fine-tuning Projection Layer \textit{versus} the Entire Multimodal LLM} \label{sec:comparison}

We are interested in exploring two potential fine-tuning strategies that could help an MLLM in gaining domain-specific visual capabilities. The first approach involves simply fine-tuning the vision-to-language projection, e.g., a simple two-layer MLP with $\sim$20M parameters. The second approach involves training the entire MLLM -- i.e., the projection layer + the LLM with $\sim$7B parameters. We conduct all our experiments with the LLaVA-1.5 model~\cite{liu2023improved}, which uses the LLaMA-2-7B~\cite{touvron2023llama} as the LLM backbone, as it is a strong representative of open-source state-of-the-art multimodal LLMs~\cite{ge2023mllm, liu2023hallusionbench, yu2023mm}.

\begin{table*}[!t]
    \centering
    \resizebox{1.0\linewidth}{!}{%
    \begin{tabular}{l cc cc cc cc cc cc cc}
    \toprule
    {\sc \textbf{Models/Variants}} & \multicolumn{2}{c}{\sc Agriculture} & \multicolumn{2}{c}{\sc Textures} & \multicolumn{2}{c}{\sc Dermatology} & \multicolumn{2}{c}{\sc Humanitarian} \\
    \cmidrule(lr){2-3} \cmidrule(lr){4-5} \cmidrule(lr){6-7} \cmidrule(lr){8-9}
     & $F_1$ & Acc. & $F_1$ & Acc. & $F_1$ & Acc. & $F_1$ & Acc.\\
    \midrule
    Random (Uniform) & 0.0309 & 0.0339 & 0.0214 & 0.0218 & 0.0451 & 0.0483 & 0.2425 & 0.2664 \\
    \midrule
    CLIP (Zero-shot; LLaVA-1.5's vision encoder) & 0.4165 & 0.4492 & 0.4582 & 0.4984 & 0.1783 & 0.2401 & 0.4139 & 0.4718 \\
    \midrule
    LLaVA-1.5 (Zero-shot) & 0.1064 & 0.1255 & 0.1882 & 0.2138 & 0.0658 & 0.0672 & 0.5169 & 0.5678 \\
    \midrule
    LLaVA-1.5 (FT-Proj with labels) & 0.2221 & 0.2478 & 0.4505 & 0.4654 & 0.2932 & 0.3403 & 0.6227 & 0.7151 \\
    \midrule
    LLaVA-1.5 (FT-E2E with labels) & 0.5984 & 0.6525 & 0.7446 & 0.7496 & 0.4947 & 0.5464 & 0.7950 & 0.8554 \\
    \bottomrule
    \end{tabular}%
    }
    \caption{{\textbf{Performance on domain-specific image classification datasets.} Fine-tuning LLaVA-1.5 end-to-end leads to the best domain-specific performance, while only fine-tuning the projection leads to a notable gain over LLaVA's zero-shot capabilities across all the datasets. It is worth noting that CLIP's zero-shot performance, which is the pre-projection image representation that LLaVA uses, is notably better than LLaVA's zero-shot performance. All the values are averaged over $5$ experimental runs with different random seeds; the $\sigma$ is $< 1\%$ for all values.} 
    }
    \label{tab:classification_results}
\end{table*}

\vspace{0.05in}
\noindent\textbf{Setting 1: Only fine-tuning the projection layer.}
LLaVA-1.5 involves pre-training the cross-modal projection layers to align image features with the pre-trained LLM's token embeddings by maximizing the next-token prediction likelihood of the MLLM.
Let $\mathbf{X}_a$ denotes the ground-truth output corresponding to the question $\mathbf{X}_q$ regarding the image encoding $\mathbf{X}_v$, which is obtained from the frozen vision-encoder of CLIP~\cite{radford2021learning}. The projection layer, parameterized by $\phi$, is trained to elicit the correct response from the frozen LLM, token-by-token while using the projected image-encoding $\mathbf{H}_v = \phi(\mathbf{X}_v)$, and considering previous tokens of the ground-truth answer. See Figure \ref{fig:appendix-formulation} (Appendix) for a pictorial illustration of the formulation. Since our focus is to perform domain-specific image classification using MLLMs, we consider $\mathbf{X}_a = \text{\texttt{<label>}}$ for a given image and construct $\mathbf{X}_q$ as: 
\vspace{-1mm}
\begin{tcolorbox}[colback=gray!10,
                  colframe=gray, 
                  arc=1mm, 
                  boxsep=0.1mm, 
                  boxrule=0.2pt] 
\small{Classify this image into one of the following categories relating to \texttt{<task>}: \texttt{<classes\_string>}. Only output a single final classification label and NOTHING ELSE.}
\end{tcolorbox}
\vspace{-1mm}
\noindent For each example, we randomly shuffle the order of classes inside \texttt{<classes\_string>} to avoid any position bias. We fine-tune the projection layers of the LLaVA-1.5 model for $1$ epoch using the default hyper-parameters \citep{liu2023improved}. During inference, we perform zero-shot classification using the same prompt above for the MLLM with the updated projection.

\vspace{0.05in}
\noindent\textbf{Setting 2: Fine-tuning the MLLM end-to-end.}
Alternatively, we fine-tune all the MLLM parameters, i.e., the projection layers and the LLM parameters concurrently by maximizing the next token-prediction likelihood of the MLLM. In other words, we update both $\phi$ and $\theta$, where $\theta$ denotes the LLM paramters. We use the same strategy to construct $\mathbf{X}_a$ and $\mathbf{X}_q$ as in the previous setting. Again, we fine-tune the LLaVA-1.5 model for $1$ epoch using the default hyper-parameters. Similar to the above setting, after training the MLLM, we perform zero-shot domain-specific image classification using the $\mathbf{X}_q$ constructed above.

We fine-tune the MLLM using these $2$ strategies for each of the $4$ datasets from different domains.

\vspace{0.01in}
\noindent \textbf{Image datasets.} The $4$ image classification datasets correspond to the following tasks: leaf disease classification, visual texture detection, skin disease identification, and humanitarian category classification. Figure \ref{fig:data-illustration} (Appendix) provides an illustration of the datasets under consideration.\\
\textit{(i) Agriculture}: To enable scalable and early plant disease detection,~\citet{singh2020plantdoc} curated PlantDoc. The dataset comprises 2,598 images categorized into 17 classes of leaf diseases. \\
\textit{(ii) Textures}: With an aim to evaluate whether visual models can identify human-centric attributes like texture beyond detecting or describing objects/scenes,~\citet{cimpoi14describing} curated 5,640 images categorized into 47 texture-related classes (like polka-dotted, wrinkled, and honeycombed). \\
\textit{(iii) Dermatology}: We consider the DermNet dataset~\citep{rimi2020derm}, which comprises 19,561 images categorized into 23 types of skin diseases like Acne, Melanoma, Seborrheic Keratoses, etc. \\ 
\textit{(iv) Humanitarian}: To aid development of computational methods that can help humanitarian organizations process images posted on social platforms during crises, ~\citet{crisismmd2018icwsm} and \citet{multimodalbaseline2020} curated the CrisisMMD dataset, which comprises 10,461 images categorized into $4$ different categories. This dataset comprises images that are the closest to natural/internet images. 

\vspace{0.01in}
\noindent\textbf{Domain-specific classification performance.} Table \ref{tab:classification_results} shows the image classification performance (macro-averaged $F_1$ scores and accuracy) of the MLLMs under various settings. For reference, we include zero-shot classification performance of CLIP\footnote{\url{https://huggingface.co/openai/clip-vit-large-patch14-336}~\cite{wolf2019huggingface}}, which is the visual encoder of the LLaVA-1.5 model (see Appendix \ref{sec:zs_clip} for details). First, it is worth noting that the zero-shot performance of the original LLaVA-1.5 model is notably worse than CLIP's zero-shot performance. This indicates that while domain-specific image attributes are present in the pre-projection image embeddings that are obtained from a frozen vision encoder (i.e., $\mathbf{X}_v$), they are not being used by the MLLM parameters. This can be attributed to the corpus used to train MLLMs like LLaVA, which comprises natural images. Second,
clearly, the results show that finetuning indeed improves performance on domain-specific classification, with significant improvements made when fine-tuning the entire MLLM (`FT-E2E') as opposed to only the projection layer (`FT-Proj'). The greater effectiveness of the FT-E2E can be attributed to greater representational space ($\sim7B$) over FT-Proj ($\sim20M$). With these observations, next, we focus on investigating the role of projection in capturing domain-specific image attributes.

\section{Role of Projection in Learning Domain-Specific Image Attributes}
\label{sec:hook_and_extract}
Following up on results in Table \ref{tab:classification_results}, we ask: \textit{does the  projection learn to model the domain-specific image attributes on fine-tuning the MLLM?}

\vspace{0.05in}
\noindent\textbf{Estimating post-projection richness.} To answer the above question, we develop a reliable-yet-simple way to estimate domain-specific richness of the projected image representation, i.e., the post-projection representation, denoted by $\mathbf{H}_v = \phi(\mathbf{X}_v)$. We do this by training an independent multilayer perceptron (MLP) to perform the image classification task using $\mathbf{H}_v$ as the image representation. This classifier helps estimate the extent of domain-specific information (or expressive power~\cite{bengio2012unsupervised}) that can be extracted from the input, in this case the post-projection image representation $\mathbf{H}_v$. In other words, a better classification performance by this MLP will denote relative domain-specific richness of the post-projection embeddings used for training, and vice versa. We train one MLP each using the post-projection representation $\mathbf{H}_v$ obtained from the following three settings: (i) original LLaVA-1.5, (ii) LLaVA-1.5 with fine-tuned projection, and (ii) LLaVA-1.5 with end-to-end fine-tuning, while keeping the architecture of the MLP the same for consistent comparison. We provide the additional details, including architecture and training hyper-parameters, in Appendix \ref{sec:richness_mlp}.

\begin{table}[!t]
\centering
\resizebox{\linewidth}{!}{
\begin{tabular}{lccc}
\toprule
\textbf{Task} & \textbf{Setting} & \thead{\textbf{Post-proj MLP}\\ (LLaVA-1.5; $F_1$)} & \thead{\textbf{MLLM}\\ (LLaVA-1.5; $F_1$)} \\
\midrule
Agriculture & Original & 0.5701 \textcolor{gray}{\small{(-----------)}} & 0.1064 \textcolor{gray}{\small{(-------------)}} \\
            & FT-Proj & 0.4134 \textcolor{dec}{\small{(-27.49\%)}} & 0.2221 \textcolor{inc}{\small{(+108.74\%)}} \\
            & FT-E2E & 0.5346 \textcolor{dec}{\small{(-06.22\%)}} & 0.5984 \textcolor{inc}{\small{(+462.41\%)}} \\
\midrule
Textures & Original & 0.6401 \textcolor{gray}{\small{(-----------)}} & 0.1882 \textcolor{gray}{\small{(-------------)}} \\
         & FT-Proj & 0.4736 \textcolor{dec}{\small{(-26.01\%)}} & 0.4505 \textcolor{inc}{\small{(+139.37\%)}} \\
         & FT-E2E & 0.6212 \textcolor{dec}{\small{(-02.95\%)}} & 0.7446 \textcolor{inc}{\small{(+295.64\%)}} \\
\midrule
Dermatology & Original & 0.3105 \textcolor{gray}{\small{(-----------)}} & 0.0658 \textcolor{gray}{\small{(-------------)}} \\
            & FT-Proj & 0.2182 \textcolor{dec}{\small{(-29.72\%)}} & 0.2932 \textcolor{inc}{\small{(+345.59\%)}} \\
            & FT-E2E & 0.2525 \textcolor{dec}{\small{(-18.67\%)}} & 0.4947 \textcolor{inc}{\small{(+651.82\%)}}  \\
\midrule
Humanitarian & Original & 0.7498 \textcolor{gray}{\small{(-----------)}} & 0.5169 \textcolor{gray}{\small{(-------------)}} \\
             & FT-Proj & 0.6025 \textcolor{dec}{\small{(-19.64\%)}} & 0.6227 \textcolor{inc}{\small{(+020.47\%)}} \\
             & FT-E2E & 0.7238 \textcolor{dec}{\small{(-03.46\%)}} & 0.7950 \textcolor{inc}{\small{(+053.80\%)}} \\
\bottomrule
\end{tabular}%
}
\caption{{\textbf{Estimating the domain-specific richness of the post-projection image representation using an independent MLP.} Compared to the original LLaVA-1.5 model, both fine-tuning strategies lead to worsened domain-specific richness of the post-projection image representation (second-last column), while the MLLM performance (last column) improves consistently. This implies that the domain-specific attributes are identified in the LLM, even when the LLM parameters are kept frozen as the projection is updated (i.e., `FT-Proj').}  
}
\label{tab:proj_richness}
\end{table}

\vspace{0.01in}
\noindent\textbf{Comparing domain-specific richness of post-projection representation across different settings.} Table \ref{tab:proj_richness} shows: \textit{(a)} the domain-specific richness of post-projection representation $\mathbf{H_v}$ (`Post-proj MLP'), and \textit{(b)} the corresponding MLLM performance (`MLLM'), across the three settings mentioned above (i.e., `Original', `FT-Proj', and `FT-E2E'). We report the macro-averaged $F_1$ score on the test set of the respective dataset for both (a) and (b). There are two key trends in Table \ref{tab:proj_richness}:  \textit{first}, when the `Original' LLaVA-1.5 model's projection layer is fine-tuned (`FT-Proj'), the domain-specific richness of the post-projection representation diminishes, while a boost in the MLLM performance is observed. Similarly, \textit{second}, with end-to-end fine-tuning of LLaVA-1.5 (`FT-E2E'), the domain-specific richness of the post-projection representation worsens while the MLLM performance boosts notably. These two trends are consistent across all the datasets considered in our study. 

\vspace{0.01in}
\noindent\textbf{Domain-specific attributes are identified within the LLM.} The two trends observed above reinforce the idea that as the MLLM gains previously-absent domain-specific image classification abilities via fine-tuning, the contribution of the projection layer in identifying relevant image attributes declines. Let us consider the two fine-tuning settings separately. In the first setting, the projection layer undergoes updates to assist the \textit{frozen} LLM in more accurate label prediction, and yet captures lesser domain-specific image attributes. This indicates that the updates in projection layer merely facilitate better use of frozen LLM parameters for the domain-specific task and do not necessarily involve mapping image attributes to the frozen LLM space. 
In the second setting as well, when both the LLM parameters and projection layer undergo updates concurrently, the projection layer captures lesser domain-specific attributes, which indicates that the updates in the LLM parameters are predominantly responsible for the acquired domain-specific image classification capabilities. In sum, our results indicate that the modeling of domain-specific image attributes in MLLMs is done by the LLM parameters, whether they are kept frozen or undergo updates. 

\section{Discussion and Implications}
\label{sec:discussion}
Existing literature on interpretability of neural networks has discussed the notion of ``multimodal neurons'' -- neurons that trigger in response to particular concepts spanning disparate modalities~\cite{goh2021multimodal, schwettmann2023multimodal, pan2023finding}. For instance, ~\citet{goh2021multimodal} demonstrate that in the CLIP model, a single neuron could respond to the photographs, drawings, or images that relate to, let's say `spiderman,' even though the input image may differ in terms of low-level visual attributes like color, edges, and corners. Similarly, ~\citet{schwettmann2023multimodal} show that a specific neurons within a \textit{frozen} text-only Transformer are responsible for detecting visual concepts, let's say like `horses,' in the input images that are projected to align with the text-only transformer. Our study adds to this literature by showing that even the acquired abilities to detect visual attributes in an MLLM are reliant on the LLM parameters. Notably, when the LLM parameters are frozen, the cross-modal projection layer adapts to facilitate detection of visual attibutes in the LLM without extracting domain-specific attributes. In other words, when the LLM is frozen and the projection is fine-tuned, the projection parameters are updated to leverage the pre-existing domain-specific knowledge in the LLM parameters. In the future, we aim to interpret the layer- \& neuron-level contributions in LLMs towards acquired multimodal reasoning.

\section{Limitations and Broader Perspective}
\textit{Limitations and future work}: Our current work focuses on a representative cross-modal projection scheme (multilayer perceptron) in a state-of-the-art MLLM (LLaVA-1.5). Other open-source MLLMs have considered other projection schemes like a trainable linear layer (LLaVa-1; ~\citet{liu2023visual}), gated cross-attention (Flamingo; ~\citet{alayrac2022flamingo}), and Q-Former (InstructBLIP; ~\citet{Dai2023InstructBLIPTG}). Future work could extend the current study to other projection schemes and models. Beyond the adopted strategy of estimating the post-projection richness of image representations using an independent classifier, future work could also probe the MLLM using concept bottleneck methods~\cite{koh2020concept}, or analyze mutual information between representations~\cite{bachman2019learning}. Finally, while outside the scope of the current work, a holistic evaluation of the MLLM should focus on domain-specific capabilities as well as the general purpose capabilities. 

\vspace{0.01in}
\noindent\textit{Broader social impact}: The authors do not foresee
any negative social impacts of this specific work. However, we acknowledge that existing LLMs and MLLMs demonstrate different forms of biases~\cite{wan2023kelly, nwatu2023bridging} that could be inherited in domain-specific variants. In line with the ongoing effort towards mitigating social biases in deep neural networks, future efforts that aim to interface modality-specific reasoning with LLMs, should consider the additional biases that LLMs may introduce on top of the modality-specific networks. 

\vspace{0.01in}
\noindent\textit{Datasets and code}: The datasets used in this study are publicly available and were curated by previous research. We abide
by their terms of use. We release the code for our experiments to aid reproducibility and enable future research on this topic: \url{https://github.com/claws-lab/projection-in-MLLMs}

\section{Acknowledgements}
This research/material is based upon work supported in part by
NSF grants CNS-2154118, ITE-2137724, ITE-2230692, CNS2239879, Defense Advanced Research Projects Agency (DARPA) under Agreement No. HR00112290102 (subcontract No. PO70745), CDC, and funding from Microsoft. Any opinions, findings, and conclusions or recommendations expressed in this material are those of the author(s) and do not necessarily reflect the position or policy of DARPA, DoD, SRI International, CDC, NSF, and no official endorsement should be inferred. Gaurav is partially supported by the JP Morgan AI Research PhD Fellowship and the Snap Research Fellowship. We thank the members of the CLAWS Lab for their helpful feedback.

\bibliography{anthology,custom}
\bibliographystyle{acl_natbib}

\appendix

\section{Appendix}

\subsection{Zero-Shot Classification Using CLIP}
\label{sec:zs_clip}
We perform zero-shot classification using the CLIP model (\texttt{clip-vit-large-patch14-336};~\citet{}), which is the same as the vision encoder used for obtaining pre-projection representation of the input image (i.e., $\mathbf{X}_v$) by the LLaVA-1.5 model. The CLIP model embeds both image and text data into a common space using a contrastive learning objective. We use the pre-trained model to compute the cosine similarity between the image representations and the representation of the dataset-specific label strings obtained from the textual backbone of CLIP. Following this, we consider the most similar label string to be the predicted label for the given image, and compute classification metrics on the test set to quantify CLIP's zero-shot performance.  

\begin{figure}[!t]
    \centering
    \includegraphics[width=0.45\textwidth]{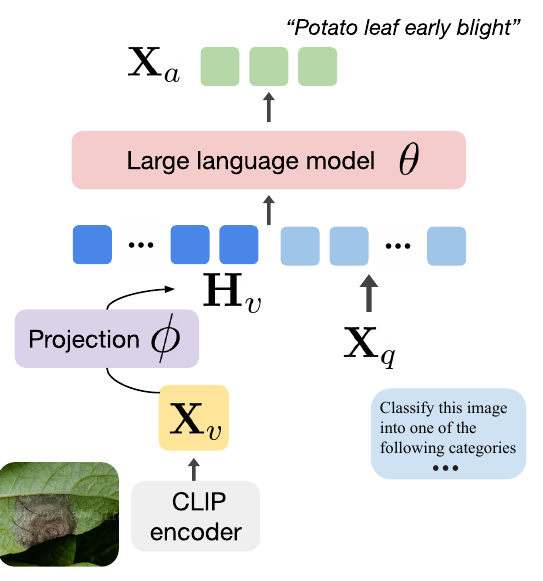}
    \caption{\textbf{Architecture of the MLLM} considered in this study. $\phi$ and $\theta$ denote tunable parameters of the projection and the large language model, respectively.}
    \label{fig:appendix-formulation}
\end{figure}

\subsection{Multilayer Perceptron for Estimating Post-Projection Richness}
\label{sec:richness_mlp}
We train a multilayer perceptron for estimating the domain-specific richness of the post-projection image representation (i.e., $\textbf{H}_v$). The MLP takes the tokens corresponding to the image as input and learns to perform the classification task using the examples from the standard train set. Architecturally, the MLP comprises a token-level average pooling step to obtain the image representation, followed by subsequent layers, and eventually the output layer of size equivalent to the number of classes in the dataset. We use ReLU activation~\cite{agarap2018deep} to induce non-linearity. We keep the architecture of this MLP fixed across all the settings to control for the number of learnable parameters and the representational power of the neural network, therefore allowing us to estimate the richness of the input embeddings with respect to the target task. Each model is trained with a batch size of 128. We use Adam optimizer ~\cite{kingma2014adam} with a learning rate initialized at $10^{-4}$ and adopt early stopping based on the loss values to avoid overfitting. As a sanity check, we note that an MLP trained using our setup on the post-projection embeddings obtained from the original LLaVA-1.5 model for the {\sc \small Humanitarian} task (a natural images dataset), achieves close to the state-of-the-art performance reported on this task~\cite{crisismmd2018icwsm}. This indicates that our setup enables a reliable estimate of the richness/expressive power of the post-projection representations.

\subsection{Performance of Image-only Models}
\label{sec:image_only}
As reference to the performance of MLLM's domain-specific capabilities (before and after fine-tuning), we include the performance of simple image-only classification models. We use the 1024-dimensional image embeddings obtained from a pre-trained CLIP model (\texttt{clip-vit-large-patch14-336}) and train a multilayer perceptron with layers of size ($1024$ (input layer), $2000$, $3600$, $1024$, $600$, $256$, \# of classes (output layer)). We use the same design choices as used for training the MLPs described in Sec. \ref{sec:richness_mlp}, and evaluate the models on respective test sets of the dataset. The results are presented in Table \ref{tab:app_image_only}. Although it is not the primary focus of this work, it is interesting to note that for the domain-specific tasks -- i.e., all the $3$ tasks except {\sc \small Humanitarian} the MLP (with $\sim20M$ parameters) performs better than the fine-tuned MLLM (with $\sim7B$ parameters). Both the model use CLIP embeddings as input representation of the image and are fine-tuned with the same amount of labeled data.

\begin{table}[!t]
\centering
\resizebox{0.75\linewidth}{!}{
\begin{tabular}{lcc}
\toprule
\textbf{Task} & {$F_1$ score} & Acc. \\
\midrule
Agriculture &  0.6991  & 0.7118 \\
Textures & 0.7644 & 0.7638 \\
Dermatology &  0.6046 & 0.6492 \\
Humanitarian &  0.7506  & 0.8238 \\
\bottomrule
\end{tabular}%
}
\caption{{\textbf{Classification performance of MLP-based image-only classifiers.} A simple MLP performs better on $3$ out of $4$ tasks than the fine-tuned multimodal LLM; see Table \ref{tab:classification_results} for MLLM results.}
}
\label{tab:app_image_only}
\end{table}

\begin{figure}[!t]
    \centering
    \includegraphics[width=\linewidth]{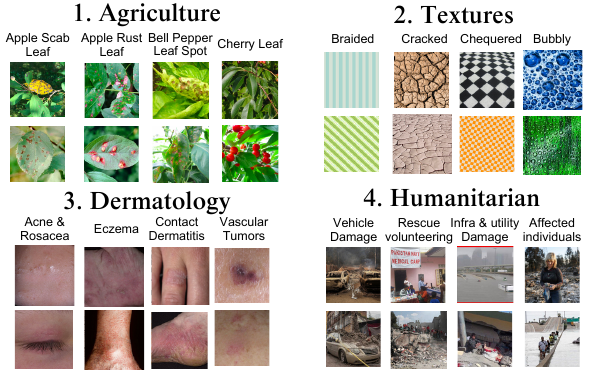}
    \caption{\textbf{Illustration of the $4$ domain-specific image classification datasets} used in this study. The datasets are from diverse domains; for brevity we only show some of the representative labels from each of the datasets. Images best viewed with zoom.}
    \label{fig:data-illustration}
\end{figure}

\subsection{Compute Resources}
\label{sec:compute}
All the experiments discussed in this study were conducted using two NVIDIA A100 GPUs (80 GB). Each fine-tuning run of the MLLM took about 1 hour requiring both the GPUs, with additional time for inference; multiple inference runs could be carried over a single GPU. The training and evaluation of the MLPs took less than 20 minutes each. Each run of zero-shot evaluation of CLIP was done on a single GPU in less than 15 minutes.

\balance

\end{document}